\documentclass[sigconf]{acmart}
\usepackage{graphicx, float}
\AtBeginDocument{%
  \providecommand\BibTeX{{%
    \normalfont B\kern-0.5em{\scshape i\kern-0.25em b}\kern-0.8em\TeX}}}

\acmConference{}
\acmYear{}
\acmISBN{}
\acmDOI{}
\acmPrice{}
\renewcommand\footnotetextcopyrightpermission[1]{} % Removes footnote with conference information in first column
\begin{document}

\title{On a Scale-Invariant Approach to Bundle Recommendations in Candy Crush Saga}

\author{Styliani Katsarou}
\authornotemark[1]
\email{stella.katsarou@king.com}
\affiliation{%
 \institution{King}
 \city{Stockholm}
 \country{Sweden}}

\author{Francesca Carminati}
\authornote{Both authors contributed equally to this research.}
 \email{francesca.carminati@king.com}
\affiliation{%
 \institution{King}
 \city{Stockholm}
 \country{Sweden}}

 \author{Martin Dlask}
\email{martin.dlask@king.com}
\affiliation{%
 \institution{King}
 \city{Stockholm}
 \country{Sweden}}

\author{Marta Braojos}
\email{marta.braojos@king.com}
\affiliation{%
 \institution{King}
 \city{Stockholm}
 \country{Sweden}}

\author{Lavena Patra}
\email{lavena.patra@king.com}
\affiliation{%
 \institution{King}
 \city{Stockholm}
 \country{Sweden}}

\author{Richard Perkins}
\email{richard.perkins@king.com}
\affiliation{%
 \institution{King}
 \city{Stockholm}
 \country{Sweden}}

\author{Carlos Garcia Ling}
\email{carlos.garcia2@king.com}
\affiliation{%
 \institution{King}
 \city{Stockholm}
 \country{Sweden}}

 \author{Maria Paskevich}
\email{maria.paskevich@king.com}
\affiliation{%
 \institution{King}
 \city{Stockholm}
 \country{Sweden}}

\renewcommand{\shortauthors}{}

\begin{abstract}
A good understanding of player preferences is crucial for increasing content relevancy, especially in mobile games. This paper illustrates the use of attentive models for producing item recommendations in a mobile game scenario. The methodology comprises a combination of supervised and unsupervised approaches to create user-level recommendations while introducing a novel scale-invariant approach to the prediction. The methodology is subsequently applied to a bundle recommendation in Candy Crush Saga. The strategy of deployment, maintenance, and monitoring of ML models that are scaled up to serve millions of users is presented, along with the best practices and design patterns adopted to minimize technical debt typical of  ML systems. The recommendation approach is evaluated both offline and online, with a focus on understanding the increase in engagement, click- and take rates, novelty effects, recommendation diversity, and the impact of degenerate feedback loops. We have demonstrated that the recommendation enhances user engagement by 30\% concerning click rate and by more than 40\% concerning take rate. In addition, we empirically quantify the diminishing effects of recommendation accuracy on user engagement.
\end{abstract}





\maketitle

\section{Introduction}
Commonly applied in e-commerce~\cite{sarwar2000analysis, alamdari2020systematic}, recommender systems often utilize collaborative filtering~\cite{rajasekar2022design} or content-based filtering methods, hybrid approaches combining collaborative and content-based filtering \cite{mohanty2022recommender}, and deep learning-based systems\cite{zhang2019deep, fu2018novel}. 

Instead of recommending single products or pieces of content, \textit{Bundle Recommendation} is a specific type of recommendation systems where the goal is to suggest combinations or sets of items (bundles) that are likely to be of interest to the user. Bundle Recommendation (BR)~\cite{bai2019personalized,
chang2023BundleRecommendation} is a complex problem due to the following reasons: In contrast to the conventional recommendation (CR) \cite{Xiao2007,Zhang2007}, where the task involves selecting one or multiple items from a fixed, but large list, BR includes various combinations of items with arbitrary quantities. While both CR and BR lead to data sparsity problems, the number of combinations in BR is several orders of magnitude higher, deepening the problem with sparsity further on. Diversity should be ensured both across bundles as well as within the bundles, to ensure a diverse item composition.

The incorporation of recommendation systems - and specifically bundle recommendation - into online games, represents a relatively recent research area that has not yet matured in industrial settings. In this work, we introduce a new method for in-game bundle recommendation systems that has been successfully applied in the industrial context of Candy Crush Saga developed by King—an online mobile game with millions of users.

\begin{figure}
    \centering
  \includegraphics[width=0.55\linewidth]{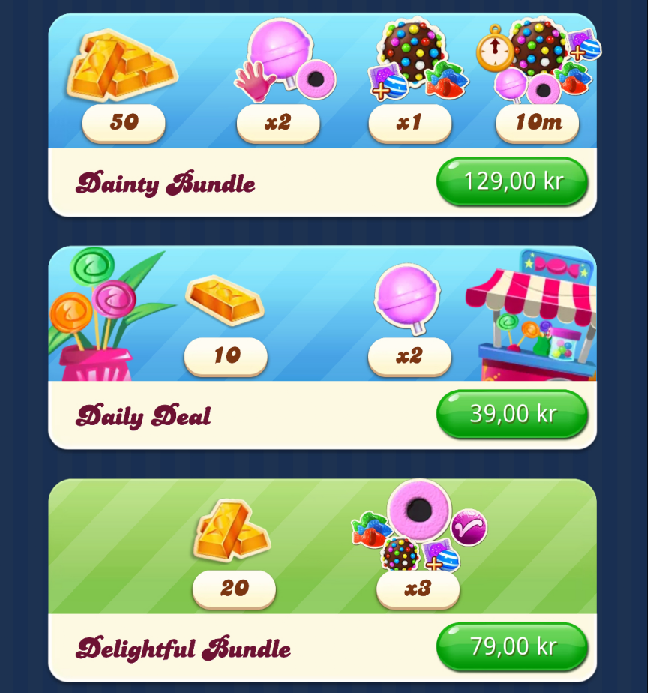}
    \caption{In-game bundle recommendation.}
    \label{fig:screenshot_game}
\end{figure}

In Candy Crush Saga, players advance through a sequential map of progressively challenging levels by solving match-3 puzzles. The pace of advancement is contingent upon the player's skill level, determined by their ability to strategically choose optimal moves, with the option of utilizing appropriate boosters and timing their usage effectively. As in other free-to-play games, players have the option to buy virtual items with real money through in-app purchases (IAPs). Users are presented with a range of bundles that consist of in-game currency and other in-game power-ups like boosters, time-limited boosters, and unlimited lives. The quantity of in-game currency and other in-bundle items can vary. An example of how bundle recommendations are presented to the users is depicted in Fig.~\ref{fig:screenshot_game}. To enhance user experience and cater to diverse playing styles, recommendation systems can help by suggesting bundles that align with players' preferences. This can increase the relevancy of the offering and contribute to higher engagement and player retention, effectively configuring the systems as value-aware. 

Furthermore, we present how our solution was deployed in the real-world in order to serve millions of users on a daily basis. Deploying machine learning systems on such a large scale poses a challenging Machine Learning Operations (MLOps) problem aggravated by the potential of accumulating technical debt. Numerous works have focused on identifying common tech debts that lead to continuous refactoring in ML systems, some of which include code duplication \cite{tang2021empirical}, non-modularity \cite{shankar2022operationalizing}, and challenges in utilizing multiple frameworks together \cite{kolltveit2022operationalizing}. While the development and initial deployment of ML systems are characterized by speed and cost-effectiveness \cite{sculley2015hidden}, long-term maintenance proves to be challenging and costly \cite{shankar2022operationalizing}. As the need to deploy and maintain ML systems efficiently and reliably increases \cite{tang2021empirical} \cite{shankar2022operationalizing}, various work focused on defining best practices for ML practitioners, such as deliberate code standardization \cite{shankar2022operationalizing}, utilizing containerization  \cite{kolltveit2022operationalizing}, proper code review, and ensuring reproducibility \cite{breck2017ml}. In this work, we describe our ML system architecture and the methods employed to mitigate technical debt in ML systems, such as minimizing production and development environment mismatch, avoiding configuration debt and inference and training skew and set up standard validation processes. These strategies are independent of the technology used and are not tied to our specific problem formulation (bundle recommendations) or system architecture, making them applicable to any ML system.

The main contributions of our paper are outlined below: 
\begin{itemize}
    \item We introduce a novel approach to bundle recommendation that employs a combination of a supervised and an unsupervised approach. The system caters player preferences, engagement and uplift with both players and the company as stakeholder. 
    \item We present the serving system, comprised of a data collection pipeline, inference and training pipelines, as well as monitoring and managed notebook pipelines. The system is integrated with our internal data warehouse and the game systems. 
    \item We share insights and patterns for building a robust and scalable ML system, with a focus on avoiding tech debt and the maintenance issues associated to it, such as production and development environment mismatch, configuration debt, inference and training skew and validation processes. 
    \item We detail the setup of online A/B experiments and explore the application of these experiments to quantify recommendation diversity and effect of feedback loops.
    \item We provide evidence of the successful deployment of our solution by quantifying post-launch performance, and we demonstrate the relationship between the offline and online performance of our model and associated engagement metrics.
\end{itemize}

\section{Related Work}

Existing recommender algorithms primarily focus on suggesting individual items based on user-item-interactions. Limited emphasis has been placed on recommending sets of items (bundle recommendation), and even less attention has been given to addressing the bundle recommendation challenge within the field of online games.

To solve the problem of bundle recommendation for suggesting booklists,~\cite{liu2014recommending} used a latent factor-based Bayesian Personalized Ranking (BPR) model, considering users' interactions with both item lists and individual items. Later, this approach was extended by~\cite{cao2017embedding} who introduced the Embedding Factorization Model (EFM), an approach that jointly models user-item and user-list interactions, incorporating Bayesian Personalized Ranking\cite{rendle2012bpr} and word2vec models\cite{mikolov2013distributed}. In~\cite{pathak2017generating}, existing bundles were suggested to users based on constituent items, and personalized new bundles were generated using a bundle-level BPR model. A graph-based approach introduced by~\cite{chang2020bundle}, unified user-item interaction, user-bundle interaction, and bundle-item affiliation into a heterogeneous graph. In~\cite{chen2019matching}, a factorized attention network was employed to aggregate item embeddings within a bundle, addressing user-bundle and user-item interactions in a multi-task manner. More recently, in~\cite{wei2023strategy}, the Bundle Graph Transformer model (BundleGT) was introduced, that utilized a token embedding layer and a hierarchical graph transformer layer to simultaneously capture strategy-aware user and bundle representations. BRUCE~\cite{avny2022bruce} is another method that adapted Transformers to the bundle recommendation problem, by leveraging the self-attention mechanism to capture latent relations between items within a bundle and users' preferences toward individual items and the entire bundle.~\cite{bai2019personalized} used a feature-aware softmax in an encoder-decoder framework and integrated masked beam search to generate high-quality and diverse bundle lists with appropriate sizes for E-commerce.~\cite{he2022bundle} introduced Bundle Multi-Round Conversational Recommendation (Bundle MCR) that extended multi-round conversational recommendation (MCR)~\cite{deng2021unified} to a bundle setting, by formulating Bundle MCR as Markov Decision Processes (MDPs) with multiple agents. Additional related work on bundle recommendation include~\cite{qi2016recommending,he2019hierarchical,garfinkel2006design,beladev2016recommender}. 

The integration of recommendation systems into online games is a relatively recent area of research that has yet to reach maturity in industrial scenarios. Previous work mostly focuses on game recommendation engines that recommend game titles based on the games users have previously played \cite{anwar2017game, sifa2015large}, or in-game single item recommendations \cite{araujo2019data,bertens2018machine,chen2018q,sifa2020matrix}. 

Specifically, the exploration of bundle recommendations for online gaming remains largely unexplored. To the best of our knowledge, only~\cite{deng2020personalized} discussed the bundle recommendation problem in on-line gaming, by framing it as a link prediction problem within a tripartite graph involving users, items, and bundles created from past interactions, and addressing it using a neural network model capable of direct learning on the structured data of the graph. Our method combines supervised learning, using an attention-based model to propose the quantities for the in-bundle items, with unsupervised learning to create the actual bundles.

\section{Methodology}

\subsection{Our Solution}\label{subsec:methodology:solution}

Suppose we have users $U=\left\{u_i \mid i=1,2, \ldots N\right\}$ and items $I=\left\{i_j \mid j=1,2, \ldots D\right\}$. Our solution comprises three sequential steps.

\paragraph{Step 1.} In the first step, we predict one D-dimensional vector per user per user $u_i$, denoted as $P_i=\left[p_{i,1}, p_{i,2}, \ldots, p_{i,D} \right]$, where each value $p_{i,j}$ represents the quantity of a bundle item $j$ purchased by the user $u_i$. 
To predict this vector, we adopt a supervised learning approach. We formulate the task as a multi-output regression problem, where the target consists of $D$ numerical values representing the quantities of each respective item purchased by the user. During training, we aim to minimize the cosine distance between true preference $P_{true}$ and prediction $P_{pred}$ as follows  
\begin{equation*}
\sum_{u_i} \mathrm{cos \_dist}(P_i^{true}, P_i^{pred}) = d_p.
\end{equation*}

Given that the targets are normalised, we use cosine similarity as our evaluation metrics as it ignores the overall scale of the predicted vectors, which is beneficial if the magnitude of the model predictions is not directly comparable to the targets. Moreover, the proportionality of the items' values in the vectors matters to us in this use-case. The cosine distance metric enables a scale-invariant comparison of the proportions of the different items present in the predicted vector and the actual label vector. 

\paragraph{Step 2.} We expect to find many similar combinations of in-game item proportions in the predictions yielded from the model in Step 1, so in this step we employ an unsupervised clustering approach to define a discrete preference space. Since the quantities of the items in a bundle are discrete, the clustering approach serves the double purpose of discretizing the problem and resolving the data sparsity. The goal here is to define a set of preference clusters, 
\begin{equation*}
    C=\left\{c_k \mid k=1,2, \ldots K \right\}.
\end{equation*} where $K$ is dependent on the later application context.
The distance from the raw prediction to the closest cluster centroid is:  
\begin{equation*}
    \mathrm{cos \_dist}(pred, clust) = d_c.
\end{equation*}

\paragraph{Step 3.} At this point we have $K$ real-valued vectors, but given that the elements of the vectors represent actual in-game products, we need to round the values so that they describe the actual quantities of the various in-game items to be shown in the bundles. 
In this step, we convert the $K$ clusters to bundles that will be recommended to our users. By the end of this step, we will have defined a set of bundles $B=\left\{b_k \mid k=1,2,...K\right\}$, where $b_k = \left (v_1,...,v_D \right ) \in \mathbb{N}^D$ with $v_j \in \mathbb{N}$ is the volumes of each item $i_j$ for every $j=1,.., D$.
The distance between the cluster centroid and the final product with rounded values is: 
\begin{equation*}
  \mathrm{cos \_dist}(clust, product)= d_o.
\end{equation*}

The error to be minimized from this whole procedure is:
\begin{equation*}
  \mathrm{cos \_dist}(true, product) <= d_p + d_c + d_o.
\end{equation*}
This 3-step process enables the segregation of the model predictions and the delivery of personalized results through bundles, providing flexibility to easily modify and market different offers.

\subsubsection{Model selection}\label{subsubsec:methodology:solution:model_selection}

Our ML model of choice for Step 1 is TabNet \cite{arik2021tabnet}. TabNet uses a structured attention mechanism to highlight important features during each decision step, which enables transparency and interpretability of the model's predictions, as well as efficient handling of sparse features. In our data, each row corresponds to a distinct user or the same user over different periods. Given the diverse user-base in terms of skill and playing style, and the dynamic nature of user playing behavior, with rapid progress and style changes over short periods, each row is unique across users and even for the same user from day to day, so not all features are expected to be relevant for every example. TabNet's capability to handle sparsity and operate on an instance-wise basis is advantageous for our use case, as it allows the model to independently determine the features to pay attention to for each example. 
Regarding TabNet hyperparameters, we primarily adhere to the default settings as provided in the PyTorch implementation\cite{pytorch_tabnet_2023}. We use a progressively decreasing learning rate schedule to enhance the stability of the model's performance. In Step 2, we chose to employ an unsupervised k-means clustering algorithm. This decision was based on its simplicity and efficiency, making it an ideal choice for scalability and speed—crucial factors when deploying for millions of predictions.

\begin{figure*}
    \includegraphics[height=150pt]{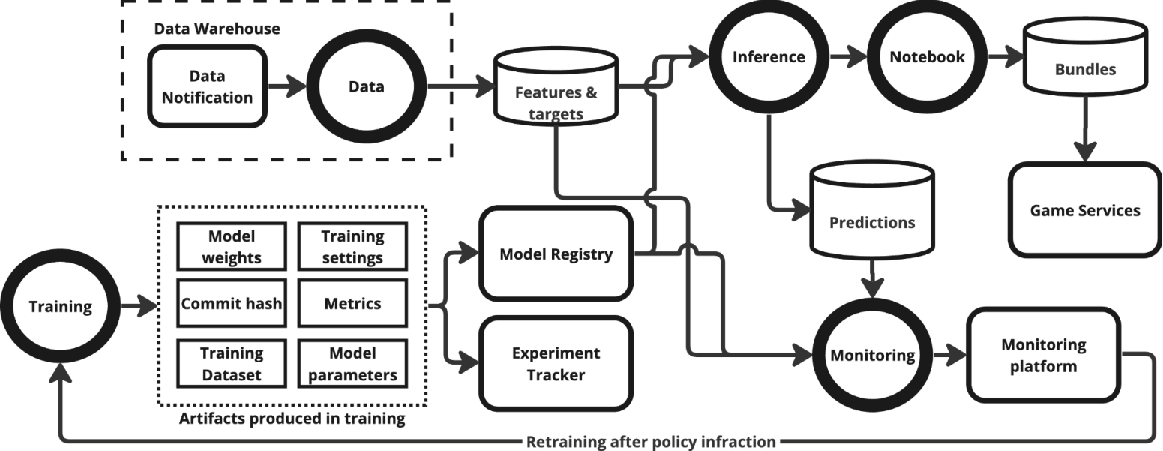}
    \caption{Pipeline Overview}
    \label{fig:PipelineOverview} 
\end{figure*}

\subsection{Model Productionization}\label{subsec:methodology:implementation}

To enhance the ease of experimenting and deploying machine learning models, King has developed a platform designed to support and automate various aspects of the ML workflow. Employing a self-service approach, the platform provides machine learning practitioners with a range of modular components and tools that streamline the modeling workflow. These resources are integrated under a unified system, akin to the previously described ML systems \cite{Markov2022} \cite{haldar2019}. Figure \ref{fig:PipelineOverview} details the structure of the training and inference pipeline. 

\subsubsection{System overview}
The system in Fig.\ref{fig:PipelineOverview} is composed of four distinct pipelines: a data pipeline for daily data extraction, a training pipeline, an inference pipeline, and a monitoring pipeline. The pipelines are depicted in the figure in bold circles. Our internal data notification system notifies our in-house pipelines orchestrator of new data availability and in turn the orchestrator triggers all pipelines.

As we only need to make daily predictions, we opt for a batch prediction system which executes all of the pipelines on a daily basis.

The initial data extraction, including the retrieval of raw model features, is handled by the data pipeline, drawing from the data warehouse. The feature transformation pipeline is configured to ensure that the feature generation process is idempotent, even in case of data backfilling. % This includes scenarios where we need to backfill data for previous periods.

The training pipeline fetches data from the data warehouse to train models and generates artifacts which are consumed by an experiment tracker and the monitoring and inference pipelines. The type of data we hold within the artifacts are model weights, parameters and metadata, the git hash of the training code, evaluation metrics, training datasets and settings (e.g. learning rate, seed, optimizer parameters, etc.). This artifact choice enables us to tackle the reproducibility challenge inherent in operationalizing ML projects \cite{sculley2015hidden} by providing all the necessary elements to recreate the same model.

The inference pipeline produces predictions for each feature for the next day and stores them in a database. %We perform batch predictions using a custom in-house component. 
After we have generated our predictions, a managed notebook pipeline is employed to map model predictions to the bundles, using the unsupervised clustering approach described in Step 2 of \ref{subsec:methodology:solution}. This enables data scientists to swiftly modify mapping logic and set up A/B experiments. In this configuration, we aggregate daily predictions into batches, which are then uploaded into the game system. The recommendation is cached in the game for the whole user session and the user is displayed the recommended bundle in the corresponding placements. To ensure reliability, the game system incorporates two fail-safe mechanisms: firstly, if there is no prediction available for a user, it defaults to the latest available prediction. Secondly, in cases where a prediction is necessary for a user but none is available, we employ a fallback bundle.

Model monitoring is essential for reliability and production-level machine learning systems \cite{breck2017ml}. Our system monitoring relies on a third-party platform, with the monitoring pipeline responsible for uploading daily predictions and features to this external platform. The monitoring pipeline is also responsible for computing the model labels, therefore it also allows us to obtain updated training data as soon as possible and use it for retraining. In addition to the standard monitoring policies to address training/serving skew, changes in feature distributions or relationship between features and labels \cite{huyen2022designing}, we track key business metrics to ensure the model's relevance to the business \cite{schroder2022monitoring}. In particular we monitor the bundles' take rate, click rate, and recommendation diversity associated with the model's usage. Furthermore, we implement feature importance monitoring to ensure that the contributions of features remain consistent during serving, fostering transparency in understanding the correlation between input data and model outcomes. Upon any monitoring policy violation, an alert prompts an investigation, followed by the model retraining; post-retraining, a detailed manual investigation informs the decision to promote the updated model to production, utilizing CI/CD pipelines for automatic deployment. 

\subsubsection{Technical Debt Prevention Strategies}
Given the inherently experimental nature of machine learning, systems designed around it can quickly develop tech debt \cite{noah2021practical}. We designed our system focusing on the ability to rapidly prototype, iterate on ideas, and deploy new models efficiently. Velocity is a crucial factor \cite{shankar2022operationalizing}, prompting us to prioritize a high experimentation pace and provide debugging environments that facilitate the swift testing of hypotheses. Simultaneously, we uphold stringent engineering standards and design patterns to ensure the reliability of our system and minimize tech debt, even during rapid iterations.  Below we describe the measures we took to prevent technical debt.
\paragraph{\textbf{Avoid development and production mismatch}} We use in-frastructure-as-code tools, leveraging standardized modules that we apply to all environments. This ensures that we have the same environments in development and production, allowing for more rigorous testing and minimizing one of the most common pain points of ML practitioners\cite{shankar2022operationalizing}.
\paragraph{\textbf{Configurations}} Configuration debt is one of the sources of tech debt in ML systems \cite{sculley2015hidden}. We expose configuration files to streamline changes across various system components. These files encompass training, data pipeline, deployment, and monitoring settings. This simplifies the comparison of changes in different iterations, as the configurations are version-controlled and incorporated into code reviews.
\paragraph{\textbf{Reducing training and inference skew}} To prevent issues related to training/serving skew, we rely on parameterized queries as the input to the model for both training and inference pipeline. This allows for a common source of truth, removing discrepancies in the data preparation. We also monitor the skew of the live data through our monitoring pipeline to ensure consistent performance. 
\paragraph{\textbf{Standard validation process}} In many ML systems, bugs stem from inconsistent definitions \cite{shankar2022operationalizing}. To avoid these issues, we have a standard validation process for all models at different stages of the ML process (offline experimentation, A/B test, business metrics). The evaluation step is embedded in the training pipeline and version-controlled. All models undergo the same A/B test setup, and once in production, we monitor the same business metrics for all of them. This ensures reliable comparisons between models and an increase in iteration speed. 
\section{Experiments}

\subsection{Existing baselines}

This section describes the results of applying our novel approach to bundle recommendations. In the offline scenario, we compare TabNet against XGBoost, while in the online setting, the comparison is made between TabNet, XGBoost as well as heuristic approach.

The heuristic approach, referred to as \textit{heuristics}, is manually designed by subject matter experts, and it is tailored to game domain knowledge rather than personalized at an individual user level. 

\subsection{Offline Experiments}
The training process is carried out in batches, with the input being a matrix  
$X=\left\{x^{(m)}\right\}^M{ }_{m=1} \in R^{M \times D}$,  where $M$ represents the number of samples in each batch, and $D$ denotes the number of input features in each sample. The input features include data on player behavior. We do not use or compare with public datasets as they do not have relevant properties and user actions that are required for our solution architecture, moreover, they cannot be used for online experiments. In our dataset, we only keep users who have been active for more than or equal to 30 days and aggregate all features by averaging them over an $N$-day period. The target consists of $D$ numerical values $\left[p_1, p_2, \ldots, p_D\right]$ representing the quantities of each respective item purchased by the user on their next active day after the $N$-day period. To simulate the production setting where users exhibit diverse activity levels, we do not aggregate the test set over a $N$-day period. Instead, we include all users regardless of the amount of active days they have had. If a user has been active for less than $N$ days, we aggregate their corresponding input features over as many days as they have been active for. 
We train two distinct models, differentiated by their respective number of aggregation days:
\begin{itemize}
    \item TabNet with $N=15$ days
    \item TabNet with $N=30$ days
\end{itemize}

We evaluate the models' performance using the mean cosine distance as our evaluation metric, as outlined in Section~\ref{subsec:methodology:solution}. A lower value of this metric indicates better model performance.

Based on the results shown in Table \ref{tab:offline_experiments_results}, we proceed with the TabNet model that uses a 30-day aggregation period.

\begin{table}[ht!]
    \centering
    \begin{tabular}{cc}
        Model & Mean Cosine Distance \\
        XGBoost baseline & 0.234 \\
        TabNet 15day  & 0.124\\
        TabNet 30day & $\mathbf{0 . 1 0 3}$ \\
    \end{tabular}
    \caption{Offline experimentation results.}
    \label{tab:offline_experiments_results}
\end{table}
\subsection{Online Experiments}
\subsubsection{A/B experimentation}
To be able to understand the online performance of our approaches to recommendation, we have tested the predictive models using A/B experiments. The A/B testing methodology allows us to compare the performance of key metrics between the treatment group and the control group in a single experiment. We denote the uplift of metric M as a percentage difference of the absolute values of M in the treatment group and control, respectively, scaled to the size of these groups. We denote the aggregate uplift in metric $M$ as $\Delta M$. 

 We define a pool of bundles as a set $\mathcal{O} = \{ O_1,O_2,...O_n\}$, where $O_i$ is a bundle. We define a random bundle $B_R$ as a discrete random variable uniformly distributed on $\mathcal{O}$, i.e. $N \sim \mathrm{U}(\mathcal{O}).$ Recommended bundle $R$ is defined as $\mathrm{argmin} \{\mathrm{cos\_dist}(O_i,x): O_i \in \mathcal{O}\} $ for model prediction $x$ and pool of bundles $\mathcal{O}$. Let $X$ be a uniformly distributed continuous random variable on $[0,100]$ and let $B_R$ be a random recommendation. The recommendation with contamination $p$\% is defined as a random variable 
$N_p =\mathrm{I}(p<X)\cdot B + I(p\geq X)\cdot P$ where I is the indicator function, i.e. $\mathrm{I}(\mathrm{true}) = 1$ and $\mathrm{I}(\mathrm{false}) = 0$ and $R$ is recommended bundle. Throughout the experiments section, we study the effect of:

\begin{itemize}
    \item Model recommendation $N_0$ ($T_0$)
    \item Contaminated recommendation  $N_{10}$ ($T_{10}$)   
    \item Contaminated recommendation $N_{30}$ ($T_{30}$)
    \item Random recommendation $B_R$ ($T_N$)
\end{itemize}

The duration of these online experiments varied because we needed to reach $80\%$ probability that the point estimate of the metric is above the intercept. For this reason, the duration varied between 14 and 42 days. The treatment groups with contaminated and random recommendations were only temporary on a tiny fraction of the daily randomized user-base to assure fairness.

\subsubsection{Metrics}
We first define the metrics that can quantify the relevancy of the recommendation.

\textbf{Click volume}: Click volume $CV(P,d)$ is the total number of clicks on product $P$ and day $d$.

\textbf{Acceptance volume}: Acceptance volume $AV(P,d)$ denotes the number of takes of product $P$ and day $d$. 

\textbf{Recommendation diversity}: Recommendation diversity $RD(d)$ is the weighted mean cosine distance with weights $AV(P,d)$ between all $P \in \mathcal{O} $ and a unit vector.

\textbf{Take rate}: Take rate $TR(d)$ is the proportion between $AV(d)$ and number of impressions on day $d$.

\textbf{Click rate}: Click rate $CR(d)$ is the proportion between $CV(d)$ and number of impressions on day $d$.

The defined metrics can help us to understand the effect of the recommendation system on the population of the users, with impact on:
\begin{enumerate}
\item comparison to existing native/heuristic approaches,
\item understanding the novelty effects associated with the content recommendation,
\item mitigating the position bias and its impact on the preference,
\item feedback loops associated with the model application.
\end{enumerate}

\subsubsection{Experiments}
We have conducted five experiments, where we tested the performance of the recommendation against the heuristic approach, which is our primary control group, but also against other treatment cases, such as random recommendation and previously developed XGBoost recommendation. The setting, including the source treatment and target treatment, is shown in Tab. \ref{tab:experiments_overview}.

\begin{table}[ht!]
    \centering
    \begin{tabular}{ccccc}
         Experiment &Source treatment & Target treatment\\
         1& random\footnotemark[1] & $T_0$\\
         2& random\footnotemark[1] & $T_0, T_{10}$,$T_{30}$ \\
         3& XGBoost rec.& $T_0$\\
         4& heuristic & $T_0$\\
         5& heuristic & $T_N$\\
    \end{tabular}
    \caption{Experimental setting overview.}
    \label{tab:experiments_overview}
\end{table}

\footnotetext[1]{Comparison to random recommendation was performed only temporarily on a tiny fraction of the daily randomized user base to assure fairness.}

\begin{figure}[ht!]
    \centering
     \includegraphics[width=230px]{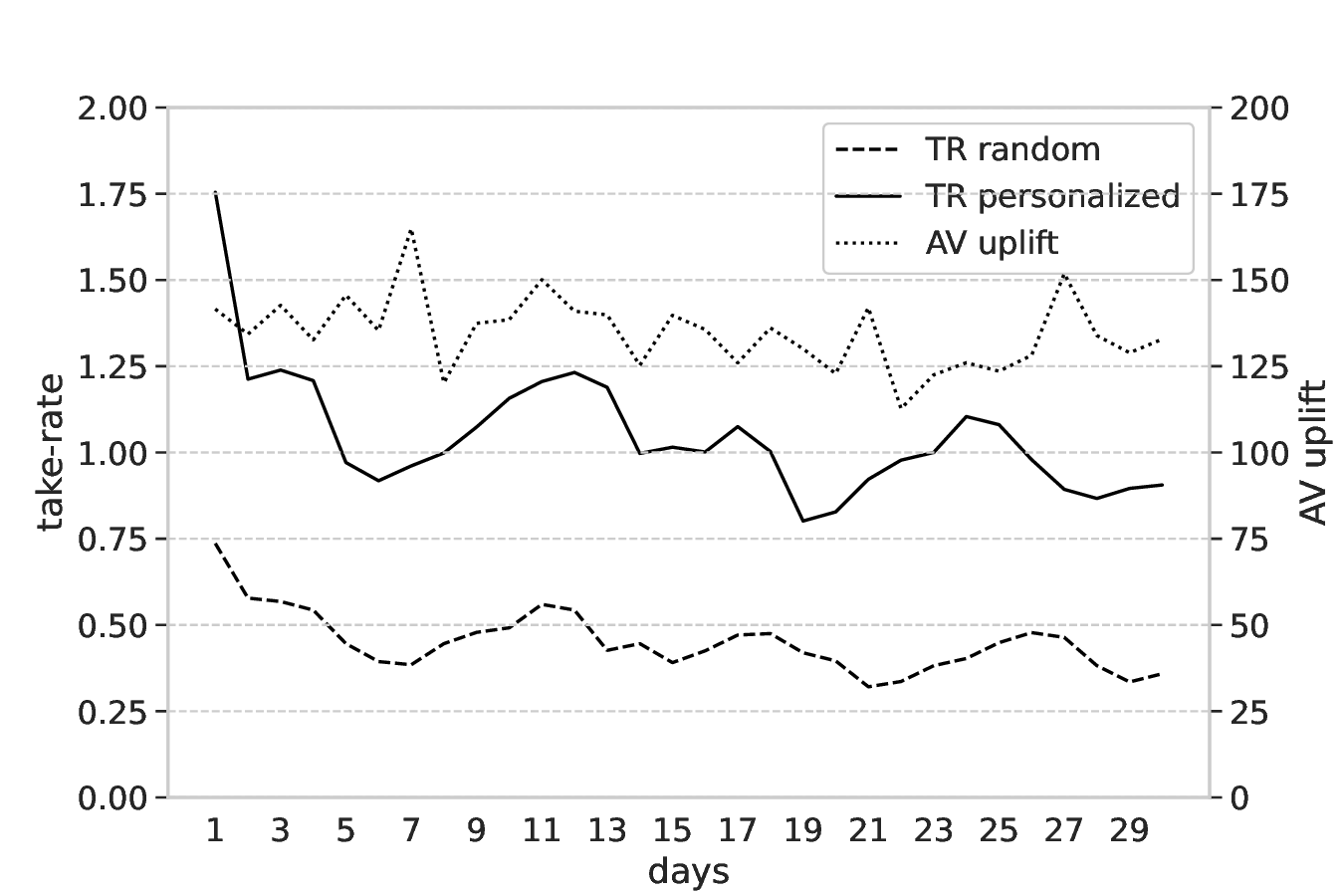}
    \caption{Novelty effect in experiment 1.}
    \label{fig: Novelty effect in experiment 2.}
\end{figure}

\paragraph{Experiment 1}
This experiment aims to study the novelty effect. The novelty effect is observed when new content is presented to users. The novelty effect is associated with an increase in uplift in engagement metrics, such as $CV$ and $AV$ increases, which often gives a false feeling of increased relevancy. Empirical experiments exhibit that the novelty effect lasts between 2 to 14 days. Especially with increasing $RD$, the novelty effect is strengthened. In this case, for comparison, we use an artificially created treatment group, where we use random bundle $B_R$ as a benchmark to simulate the novelty effect, which is tested against the recommended bundle $R$.  The appropriate length of the experiment(s) and the simulated randomized group should guarantee that the real effects will be decoupled from novelty effects. For acceptance rate, there was a novelty effect in both cases when a new bundle was introduced on the game. The experiment has shown $129.69\%$ increase in $AV$ and $+35.14\%$ increase in $CV$. The novelty effect is visualized in Fig. \ref{fig: Novelty effect in experiment 2.}. The left axis denotes the take rate of the random and personalized groups, while the right axis shows the $AV$ uplift with novelty effects removed. We can see that the recommendation provides a stable $AV$ uplift over time when the novelty effect is removed.

\paragraph{Experiment 2}
We intend to prove the relation between the model recommendation and the metrics of interest, at different levels of contamination. This will enable us to understand how the key engagement metrics deteriorate when the model is contaminated with a random recommendation. We take advantage of the artificially created treatment group with a random recommendation as a method to compare to other treatment groups $T_{0}, T_{10}, T_{30}$. The changes in key metrics are presented in Tab. \ref{tab:exp_2_results}. We can see that with increasing levels of contamination, the $AV$ and $CV$ decrease, however, at a non-linear pace. The level of contamination $p$ does not guarantee a proportional decrease in the engagement metrics. Looking at $RD$, for lower contamination, the novelty effect is likely mild, therefore locally increasing the take rate could result in a higher increase of $RD$, whereas for higher contamination, the relevance of the prediction deteriorates too much, resulting in a decline in $AV$ and resulting in smaller increase in $RD$. 

\begin{table}[ht!]
    \centering
    \begin{tabular}{cccc}
         Treatment group & $\Delta AV$ & $\Delta CV$  & $\Delta RD$ \\
         $T_0$    &  131.41\% & 39.18\% & -1.22\%\\
         $T_{10}$ &  117.74\% & 36.64\% & 2.44\%\\
         $T_{30}$ &  91.17\% & 28.81\% & 0.61\%\\
    \end{tabular}
    \caption{Changes in engagement metrics from experiment 2.}
    \label{tab:exp_2_results}
\end{table}

\paragraph{Experiment 3}
In this experiment, we compared the performance of the TabNet model with 30-day aggregation against the XGBoost baseline model, which will also experimentally verify the performance of the advanced recommendation technique in the online space.  We observed that we've managed to increase $AV$ by 28.68\% and $CV$ by +8.28\% respectively.
While XGboost provides lower performance on the training and validation data, the model provides higher absolute $RD$. This is because its inaccuracy results in recommendations being closer to a random recommendation, similarly in the case of contamination. 

\begin{figure}[ht!]
    \centering
    \includegraphics[width=250px]{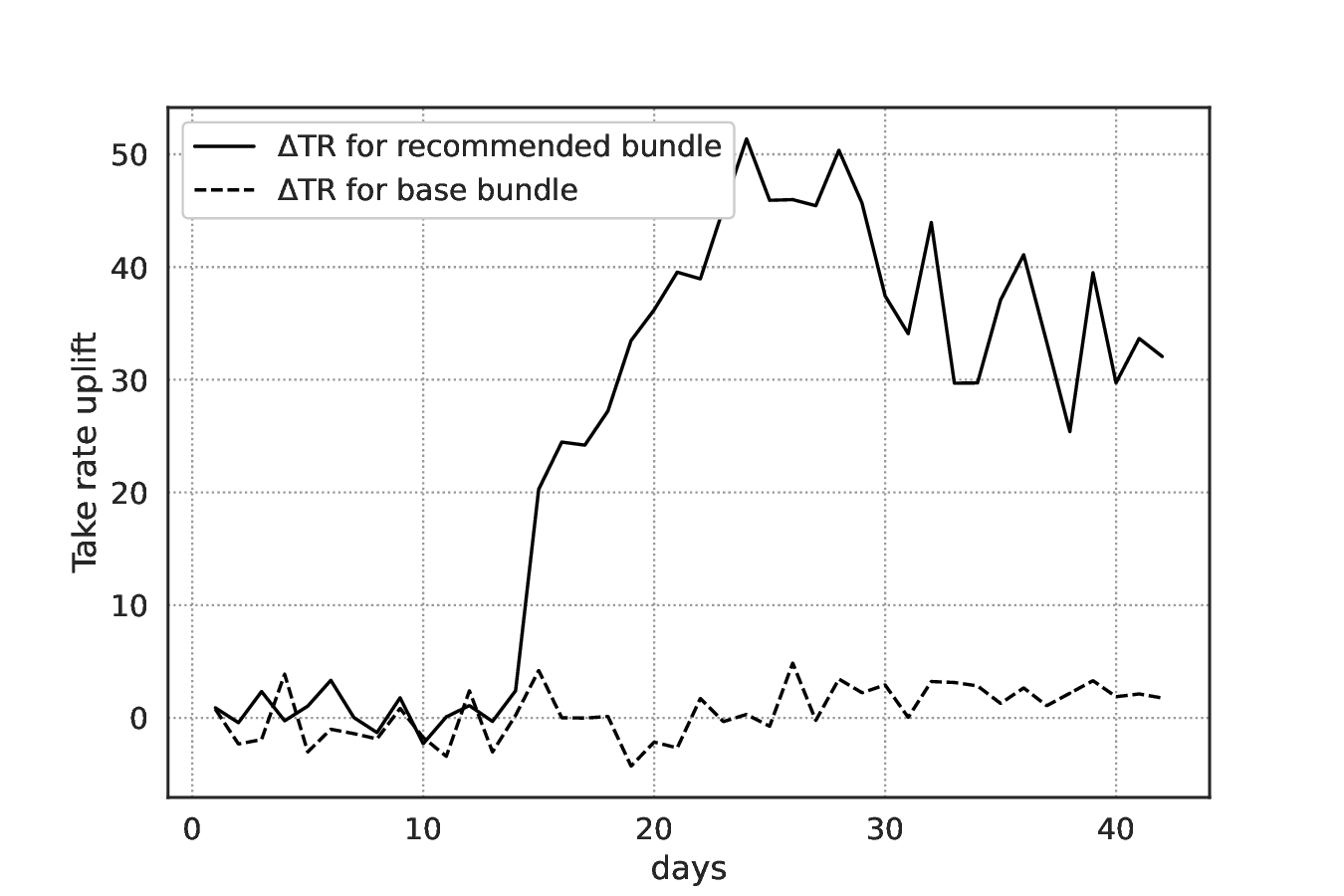}
    \caption{Take rates uplifts in experiment 4.}
    \label{fig: Take and click rates uplifts in experiment 4.}
\end{figure}

\paragraph{Experiment 4}
This experiment tests the TabNet recommendation against a heuristic recommendation. While the uplift in acceptance volume has increased significantly by 41\% and the corresponding click volume has increased by 33\%, the RD has also increased by 4.57\%. We visualize the trend in the increase of take rate of the recommended product in Fig. \ref{fig: Take and click rates uplifts in experiment 4.}. The experiment started on day 11, and while the recommended product gradually gained popularity, the novelty effect stabilized roughly 20 days after the experiment started.

\paragraph{Experiment 5}
Previously we've observed behavioral changes when comparing a model output to random groups or heuristics. To bring a better validity to the results, we also want to understand, how good are the current heuristics. Therefore, we've conducted one more experiment that compares heuristics to a treatment group with a random recommendation. This will allow us to understand how accurate the heuristics currently are compared to a random baseline. As expected, the $AV$ has significantly decreased in this case, resulting in a 40.21\% decline. The click rate, being more volatile, has decreased by 18.73\%. Interestingly, the $RD$ has increased only by +0.81\%. This is comparable to experiment 2 with higher contamination, where the novelty effect has increased the $RD$ only slightly. 

\paragraph{Experiments summary}
Results from all experiments are summarized in Tab. \ref{tab:all_experiments_results}. 
\begin{table}[ht!]
    \centering
    \begin{tabular}{cccc}
        Experiment & $\Delta AR $ & $\Delta CV$  & $\Delta RD$ \\
        1 & 129.69\% & 35.14\% & -0.89 \%\\
        3 & 28.68\% & 8.28\% & -5.35\%\\
        4 & 41.32\% & 32.59\% & 4.57\%\\
        5 & -40.21\% & -18.73\%  & 0.81\% \\
    \end{tabular}
    \caption{Changes in engagement metrics across experiments.}
    \label{tab:all_experiments_results}
\end{table}

\begin{figure}[ht!]
    \centering
    \includegraphics[width=250px]{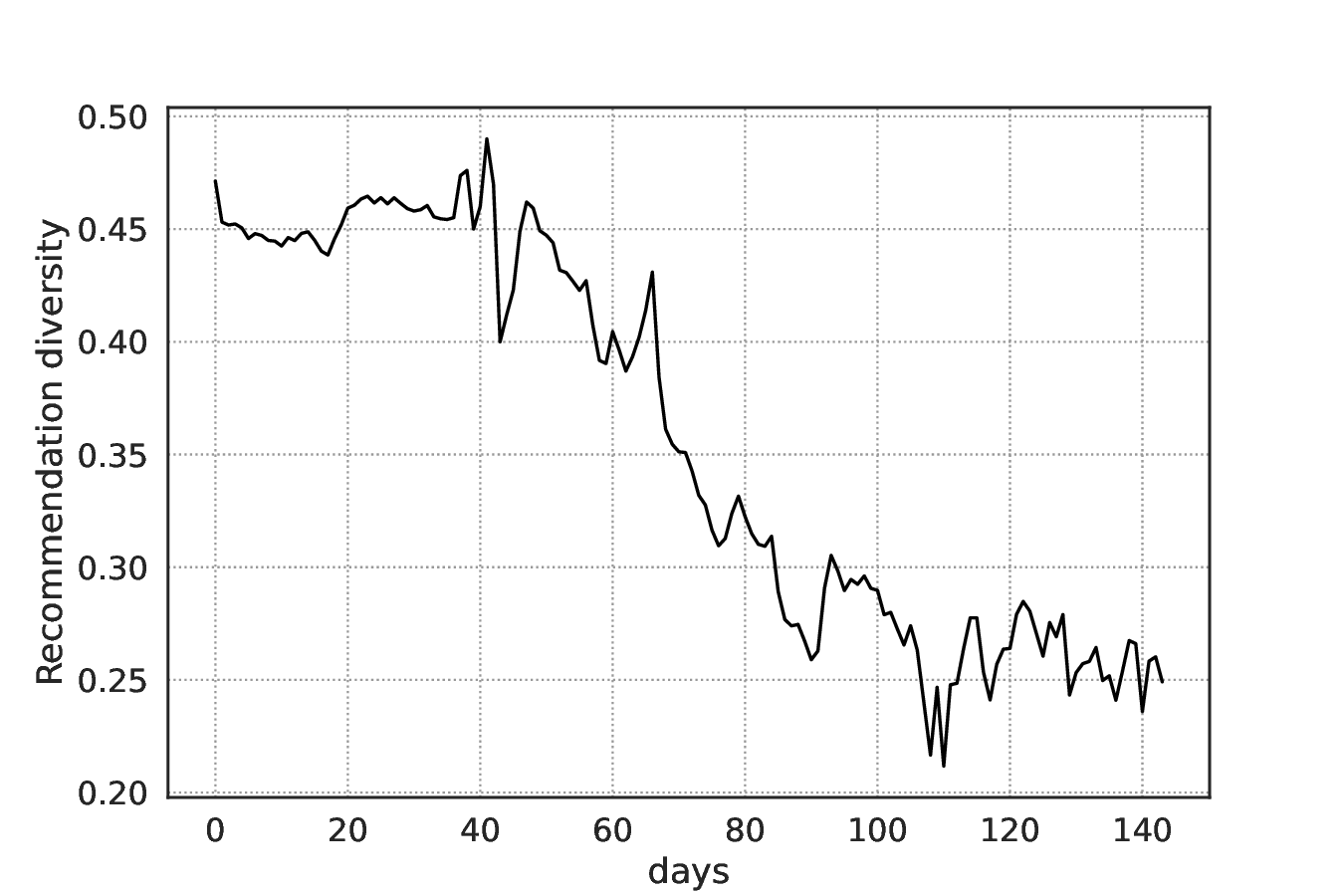}
    \caption{Recommendation diversity development over time.}
    \label{fig: Recommendation diversity development over time.}
\end{figure}

Recommendation diversity is an important metric to look into not only during an experiment but also on a global level. RD can be a proxy for discovering feedback loops and other adhere effects of our recommendation system. Once the system is trained, the RD is the highest and likely will deteriorate over time. This can be due to the effect of either model retraining or user behavior changes.
We observed a greater improvement in performance when the model was deployed for all users. This is due to the effect of a positive feedback loop: when users are presented with more relevant bundles, the bundles take rate increases. These changes in behavior lead to shifts in the data, and retraining with this new data results in a stronger feedback loop. This phenomenon is similar to "direct feedback loops" in literature \cite{sculley2015hidden} which refers to a model that directly influences the selection of its future training data, and it is more difficult to detect if they occur gradually over time. Since we monitor the shift in features and predictions, we have a measure of the impact of the feedback loop on the population. The figure \ref{fig: Recommendation diversity development over time.} visualizes the $RD$ as a dependent variable over time. We can see that it takes a few days before the $RD$ starts to deteriorate. Surprisingly, it can take up to 100 days after the diversity of the recommendation stabilizes. 

\color{black}
In our case, diverse exposure to single boosters, obtained through varied game rewards, mitigates biased recommendations associated with favoring specific products \cite{ccapan2020towards}. 
\section{Discussion}
In this paper, we present our first iteration for the approach presented in \ref{subsec:methodology:solution}. With our 3 step solution, mixing a supervised method, for the player preferences, and an unsupervised method for defining the cluster of items, we are able to simplify the bundle recommendation problem. 

\subsection{TabNet}

TabNet \cite{arik2021tabnet} has been our initial take on a Tabular Neural network for this approach due to its flexibility and interpretability. 

\subsubsection{Interpretability}
The model's architecture uses a sequential attention mechanism that dynamically identifies and prioritizes important features for each sample. Specifically, by examining attention weights,  we can ensure the model prioritizes relevant features. This helps identify and correct biases where the model might overemphasize features linked to unrelated targets, leading to more accurate, target-focused predictions.

\subsubsection{Self-supervised pre-training} 
In the TabNet model, the self-supervised pre-training refers to training initially on an unsupervised task without labels, allowing the model to discover data patterns before the supervised training step \cite{arik2021tabnet}. This method helps the model identify patterns in partially visible data and focus on important features, enhancing its ability to better identify underlying structures.

\subsubsection{Other modeling approaches}
In the current 3-step approach described in \ref{subsec:methodology:solution}, we use TabNet as the model of choice. The downside of the solution is its heavier computational load, which slows down the training pipeline. While real-time retraining is not a constraint in the current formulation, we are interested in exploring lighter models and the balance between offline performance evaluation and computation costs for the training across different architectures. XGBoost or other vanilla Neural Networks like a simple feed-forward neural network, have been considered, and initial experiments reveal comparable results in our key metric, cosine distance, during offline evaluation.

\subsection{Data and Loss Function enhancements}

Using the same data but adding some smart processing can help our models identify player preferences more accurately. In this section, we discuss a couple of approaches in this direction.

\subsubsection{Feature Aggregation}
When exploring player's preferences in historical data, it is important to capture both short-term patterns as well as more long-term in-game habits and preferences of a user. Currently, our model's 30-day input feature aggregation captures long-term player behavior but doesn't give higher weight to more recent events. Adding shorter aggregation periods could enhance model performance compared to the baseline. An alternative approach to manual feature engineering could be to build a hierarchical model where the lower-level components specialize in capturing short-term patterns, and the higher-level components aggregate information over longer time spans.

\subsubsection{The Cold Start Problem}
Given that the training data is focused on purchases and usage of in-game items,  it specifically targets users who have made a purchase on a given day. Our aim is not to predict the occurrence of a purchase but to determine the preference vector for items within the purchased bundle. As a result, the training data includes only paying users and new users, while non-paying players are excluded. This formulation introduces a cold start problem, as it necessitates at least one purchase from a user before model-based recommendations can be generated. Although this issue could be addressed by developing separate models for underrepresented player groups, it falls outside the scope of the current iteration.

\subsubsection{Loss function: imbalance between targets}
The current solution involves a multi-objective regression with 13 targets, each representing an in-game item that can be included in a bundle. The loss function is the cosine distance between the output and true label vectors, not accounting for potential weight differences between items. This can lead to overestimation if an item is over-represented in training data. Our model analysis reveals that the performance metric deteriorates by 60\% when evaluated on 12 targets, with the in-game currency included in training but excluded from evaluation. This suggests that the in-game currency alone accounts for 60\% of the cosine distance between the predictions and the true labels. To better understand the players' preferences and cater to individual play styles we may consider approaches to ensure a more balanced learning environment for the model. Examples include target-specific weighting, separate models for distinct targets, or fixed allocation for specific targets, and focusing the model on the preferences for the remaining. 
\section{Conclusions}
In this paper, we present a novel two-step approach to item recommendations in mobile games, which was applied and tested on a bundle recommendation problem in Candy Crush Saga. First, we've defined the general methodology and architecture of the solution, which was specially designed for the mobile game environment. Apart from offline validation, the architecture was also tested in several online experiments, empirically modeling the relationship between the click- and take rates and model accuracy. The robust architecture and technical debt prevention strategies allowed the system to be deployed in two in-game placements, one of them being illustrated in Figure \ref{fig:screenshot_game}. 

The novelty of this approach lies not only in the item recommendation methodology, which is subsequently applied to bundle recommendation but also in the implementation. The robust and fail-safe pipeline is designed to scale for millions of players and implements many policies that can prevent the delivery of inaccurate recommendations. We continuously monitor the system performance, both in the offline and online environment, where we focus on understanding the click change- and take rates, but also other underdeveloped metrics, such as the impact of degenerate feedback loops and a corresponding deterioration in recommendation diversity. 

The scale-invariant system defined in the methodology is an efficient tool both for the generalization of the system for other tasks and presents a responsible AI solution that makes sure fairness resulting from the generation of the recommendation is in place regardless of user's level of activity or spending.  

\section{Acknowledgements}
We're grateful for the support from King we've been given while preparing the manuscript and the openness we work in that contributes to a good level of insight sharing. In addition to the authors who have contributed hands-on to the success of this solution, we also thank other teams in King for their help and support, specifically ML Special Projects, MLOps Accelerator, Core Data, AI Labs, CCS IAP\&E and CCS Operators teams. Last but not least, we also thank Pradyumna Prasad for his significant support and contribution to the success of the project. 
\bibliographystyle{ACM-Reference-Format}
\bibliography{bibliography}
\end{document}